\documentclass[runningheads]{llncs}

\usepackage{graphicx}
\usepackage[T1]{fontenc}
\usepackage[utf8]{inputenc} %
\usepackage{fdsymbol}       %
\usepackage{newunicodechar} %
\newunicodechar{⌀}{\ensuremath{\diameter}}\usepackage{cite}

\usepackage{booktabs}
\usepackage{multirow}
\usepackage{amsmath}
\PassOptionsToPackage{hyphens}{url}\usepackage[hidelinks]{hyperref}
\usepackage{placeins}

\begin{document}

\renewcommand{\labelenumii}{\arabic{enumi}.\arabic{enumii}}

\title{Towards Lithuanian Grammatical Error Correction}
\author{Lukas Stankevičius\orcidID{0000-0003-0012-5471} \and \\
Mantas Lukoševičius\orcidID{0000-0001-7963-285X}}
\authorrunning{L. Stankevičius and M. Lukoševičius.}
\institute{Faculty of Informatics, Kaunas University of Technology, Kaunas, Lithuania}
\maketitle              %
\begin{abstract}

Everyone wants to write beautiful and correct text, yet the lack of language skills, experience, or hasty typing can result in errors. By employing the recent advances in transformer architectures, we construct a grammatical error correction model for Lithuanian, the language rich in archaic features. We compare subword and byte-level approaches and share our best trained model, achieving F$_{0.5}=0.92$, and accompanying code, in an online open-source repository.

\keywords{Natural Language Processing \and Grammatical Error Correction \and Transformer Models \and ByT5 \and Lithuanian}
\end{abstract}
\section{Introduction}

Recent advances in neural Natural Language Processing (NLP) have pushed the frontiers. In particular,  transformer-architecture-based models \cite{transformer} surpassed human performance in various NLP benchmarks such as SQuAD2.0 \cite{rajpurkar-etal-2018-know}, GLUE \cite{wang-etal-2018-glue}, and SuperGLUE \cite{NEURIPS2019_4496bf24}. This also opened new opportunities for the Grammatical Error Correction (GEC) task which we address in this work.

GEC is the task of correcting different kinds of errors in text such as spelling, punctuation, grammatical, and word choice errors. The abundance of such noise in the text can hinder not only the understanding by humans but also the performance of various downstream NLP systems. An error-free text is also more beautiful, clean, associated with a certain prestige. However, producing it may be problematic for non-native speakers, language learners, it requires additional time and effort. 

NLP state of the art for GEC still has much room to improve. As of now, the best F$_{0.5}$ scores are only up to 0.72\footnote{\url{http://nlpprogress.com/english/grammatical_error_correction.html}}. Moreover, the research is mostly focused on English and a few other popular languages. To reduce this gap  and contribute to the GEC progress, in this work, we investigate it for the Lithuanian language.

The Lithuanian language is one of the oldest living languages in the world. It has retained most of the features of the Indo-European Protolanguage, i.e., it is characterized by a very ancient linguistic structure: declensions (of nouns, adjectives, and pronouns), short and long vowels, diphthongs, etc. Lithuanian also has many similarities with Sanskrit -- the classical language of ancient India, still used today as a scholarly and liturgical language in Hinduism, Buddhism, and Jainism. Antoine Meillet (1886-1936), one of the most influential French linguists, once stated: ``Anyone wishing to hear how Indo-Europeans spoke should come and listen to a Lithuanian peasant''.

The Lithuanian language is synthetic and uses inflections to express syntactic relationships within a sentence. In other words, the relations in a sentence are expressed by word endings rather than with unbound morphemes and word order. This allows a lot of freedom in composing sentences. In contrast to agglutinative languages, which combine affixes by ``gluing'' them unchanged inside word ending, in Lithuanian inflectional categories are ``fused''. Meanwhile, prefixes, suffixes, and infixes are still used to derive words. Lithuanian verbs can be made from any onomatopoeia; phrasal verbs (e.g., go in, go out) are composed by adding the prefix to the verb. Lithuanian is unique for having 13 different participial forms of the verb \cite{uniqueLithuanian} while modern English has only 2 (present and past participles). It is estimated that 47\% of Lithuanian word forms are morphologically ambiguous \cite{rimkute2006morfologinio}, i.e., requiring context consideration to discern the meaning. All these mentioned features of the Lithuanian language make it interesting and important to analyze in the context of automatic GEC.
 
Our contributions are: 
\begin{itemize}
\item We present the first GEC system for Lithuanian language based on deep neural networks.
\item We compare sub-word and byte-level tokenization approaches for Lithuanian grammatical error correction.
\item We share all the technical details, code, and model weights for open reuse and reproducibility.
\end {itemize}

\section{Related Work}

The simplest form of GEC is spellcheck. GNU Aspel\footnote{\url{http://aspell.net/}} and Hunspell\footnote{\url{http://hunspell.github.io/}} are two widely-used open source spellcheckers. In particular, Hunspell \cite{hunspell_lt} is the only system we found for Lithuanian GEC. Such systems work by keeping a large dictionary of possible words and detecting the non-words. During detection, the nearest alternatives from the dictionary are suggested. In Hunspell's case, the dictionary is made more compact by keeping only the main word forms with transformation rules, prefixes and suffixes. Spellchek systems are compact but limited to the correction of only non-words.

The first systems for a more complex GEC were based on Statistical Machine Translation (SMT) using a noisy channel model \cite{brockett-etal-2006-correcting}. A significant contribution to GEC was the introduction of the CoNLL-2014 shared task \cite{ng-etal-2014-conll}. Multiple systems were proposed, and among them, the phrase-based SMT setup was the most promising \cite{junczys-dowmunt-grundkiewicz-2016-phrase}. Yet neural approaches started to emerge, like \cite{xie2016neural}. As such systems advanced, hybrid statistical (SMT) and Neural Machine Translation (NMT) approaches \cite{grundkiewicz-junczys-dowmunt-2018-near} took the top. Only the introduction of the Transformer model \cite{NIPS2017_3f5ee243} enabled neural approaches to supersede the statistical ones. As of now, the latter systems are claiming state-of-the-art results in GEC \cite{rothe2021a, omelianchuk-etal-2020-gector}.

Novel less-supervised approaches are also emerging. A simple language model reaching a reasonable performance with minimal annotated training data was demonstrated in \cite{bryant-briscoe-2018-language}. The proposed system used $n$-gram language model to score variations of a sentence until incremental inflections do not improve the score anymore. Such a system was again improved using transformer-based language models instead of the $n$-grams in \cite{alikaniotis-raheja-2019-unreasonable}. It turns out that such a less-supervised approach can outperform fully-supervised systems that were claiming state-of-the-art results several years ago.

Currently, the main constraint for GEC is the lack of training data. Researchers make progress by including new data sources or using automatic grammatical error generation to synthesize them. A simple language-agnostic pre-training objective was proposed in \cite{rothe2021a}. The authors automatically corrupted sentences in character level: swapping, inserting, dropping spans; token level: swapping, dropping spans; word level: lower-casing, upper-casing the first letter. A bigger model and larger dataset allowed achieving state-of-the-art GEC results for 4 languages. Authors of \cite{shah-de-melo-2020-correcting} used a small corpus of spelling errors to derive statistics for typographical error generation and generate a large parallel synthetic corpus. Another way is to use the data that the model incorrectly predicted during the training. A fluency boost learning and inference mechanism was proposed in \cite{ge2018reaching} that reuses less fluent model predictions as new inputs during subsequent epochs. Similar trends are emerging with other languages. For example, simply adding new data improved a Transformer GEC system for the Czech language \cite{naplava2022czech}. To summarize, it is important for neural GEC systems to be trained on large and high-quality corpora.

\section{Dataset}

As mentioned, a large dataset is essential for training a neural GEC solution. 
The data also has to be of the highest quality so that we can take it as a gold standard of grammatically-correct text. 
The largest publicly available general-purpose dataset for the Lithuanian language is from OSCAR \cite{OrtizSuarezSagotRomary2019} at 5\,GB of deduplicated text. However, it is obtained from a general Common Crawl\footnote{\url{https://commoncrawl.org/}} and makes trusting the grammatical correctness problematic.

To make sure that the text is of good quality, we crawled various Lithuanian websites ourselves. We manually checked how the text is structured in every webpage so that only the relevant parts: title, summary (optionally), and the main text paragraphs would be scrapped. We crawled the following types of web pages: news, literature, blogs, encyclopedias, others.

We added titles and summaries to the main text paragraphs as additional paragraphs. Finally, we split the data into paragraphs. As a result, a single paragraph became a single sample of our dataset.

\subsection{Preprocessing}

Although we performed a well-curated data scraping, there were still some artifacts in our data that had to be corrected or removed.

We had to remove some websites because of relatively high rates of spelling errors. %
This left us with a total of 34 final websites.

Some common error patterns can be easily corrected automatically. We looked at common mistakes in Lithuanian web texts \cite{tamulioniene2015budingiausios} and performed the following corrections:
\begin{enumerate}
  \item Incorrect quotation marks. In Lithuanian, the correct are ,,ABC``. Meanwhile, the English version ``ABC'' or others such as the universal "ABC" is often used instead.
  \item The lack of space. It can happen before ``m.'' and ``d.'' abbreviations. For example, the text ``1918m. vasario 16d.'' must be corrected to ``1918 m. vasario 16 d.''. Additionally, the space is often omitted after a full stop: ``ir t.t.'' and ``A.Sabonis'' must be corrected to ``ir t. t.'' and ``A. Sabonis''.
  \item An unnecessary space. The space must be omitted before most punctuation marks: ``tik darbui , visiškai pamirštant poilsį ,'' is corrected to ``tik darbui, visiškai pamirštant poilsį,''.
\end{enumerate}

We also filtered the text samples based on some statistics: %
\begin{enumerate}
  \item The sample text length should be at least 20 characters.
  \item The fraction of Lithuanian letters in a sample should be at least 0.98. This filters out text from other languages and with miscellaneous characters. In the end, we are solving a task for the Lithuanian language. We included the characters ``€₤\$\%wx'' as Lithuanian since they are used quite often.
  \item The fraction of spaces to non-spaces should be at most 0.02. This allowed us to filter out samples dominated by URL addresses.
\end{enumerate}

Lastly, we deduplicated our text samples. We shuffled the resulting 29\,312\,785 samples and took a subset of 4\,194\,304 for this work. Some statistics for the subset are depicted in Table \ref{tab:dataset}.
\begin{table}[h]
\setlength{\tabcolsep}{1.8pt}
  \caption{Dataset sizes by various tokenizations. The total dataset size is 4\,194\,304 samples.}
  \label{tab:dataset}
  \begin{tabular}{lcrc}
  \toprule
 \multirow{2}{*}{Tokenizer} &  {Sample length,} &{Tokens,}   & \multirow{2}{*}{Tokenization example}\\
 & mean $\pm$ std  & $\times 10^6$  & \\
  \midrule
Characters      & $226\pm194$ & 947&  Lietuva – graži šalis\\
ByT5 \cite{xue2021byt5} & $243\pm194$ & 1\,017 & Lietuva \symbol{92}xe2\symbol{92}x80\symbol{92}x93 gra\symbol{92}xc5\symbol{92}xbei \symbol{92}xc5\symbol{92}xa1alis\\
T5 \cite{stankevivcius2021generating} & $48\pm43$ &201 & [\_Lietuva] [\_–] [\_graži] [\_šalis]\\
mT5 \cite{xue-etal-2021-mt5}&$71\pm61$ & 298& [\_] [Lietuva] [\_–] [\_] [graž] [i] [\_šal] [is]\\
Words   & $30\pm26$ & 126 & [Lietuva] [graži] [šalis]\\
\bottomrule
\end{tabular}
\end{table}

The transformer model has a quadratic running time complexity $\mathcal{O}(n^{2})$ with respect to the sequence length $n$ (number of tokens). Usually, this is not a constraint as most text tasks are within the maximal sequence length of 512 (T5 \cite{2020t5}) sub-words or 1024 (ByT5 \cite{xue2021byt5}) bytes. Yet in our training dataset, we had longer examples that we did not wish to truncate and, hence, lose. Instead, we split these too-long sequences to the length of 2100 characters for the T5 model and 700 characters for the ByT5. After that, we proceeded with the corresponding tokenization. As a result, the exact numbers of samples and tokens differ for both models, but the initial dataset and the amount of text (see Table \ref{tab:dataset}) is the same.

For both runs, we set aside 0.05\% of the data for validation and another 0.05\% for testing.

\section{Methods}

\subsection{Generating Grammatical Errors}

We induce 3 groups of synthetic grammatical errors described below.

\subsubsection{Typographical Errors}

They are induced by modeling the way how humans mistype on the keyboard. We follow the exact same methodology as in \cite{stankevivcius2022correcting}: take mistyping statistics between each pair of characters on a QWERTY keyboard from an English dataset and apply them probabilistically to our texts. Out of the all characters considered, this way we corrupted 2\% of them; from which were: substitution, 36.1\%; deletion, 31.7\%; insertion, 17.8\%; transposition, 14.4\%.

\subsubsection{Confusing Similar Sounding Letters}

This is a very common source of spelling mistakes. We model them by defining sets of characters that sound alike, and randomly substituting a letter with one from the same set at a point of the generated error. For this, we use weighted sampling. The probabilities of the letters for substitution are proportional to how frequent they are in Lithuanian texts overall. For example, a in single set of letters ``iįy'' (where sounds differ only in their length), it is way more common to mistakenly write  ``i'' instead of ``į'', rather than ``į'' instead of ``i'', as ``i'' is much more common. Only 2\% of all such found occurrences were replaced. Groups of letters and detailed probabilities for the group members are derived from the raw (no preprocessing) subset of 2\,909\,403 samples, and are presented in Appendix \ref{appendix1}.

\subsubsection{Other Errors}

We also introduce errors in the text by the four specific rules described below. We, again, thus corrupt 2\% of the matches of the rules.

\begin{enumerate}
\item Gemination are doubled consonant letters that sound like a single one and thus is prone to be typed only once. This also applies to any consecutive letters from ``cčsšzž'' group. For example, the words ``pusseserė užsimerkė'' may be mistakenly written as ``puseserė usimerkė'' as they sound similar due to the gemination.

\item Assimilation to an adjacent letter. This is specific to any letter of ``ptksš'' being before any of the ``bdgzž'' or vice versa. For example, the words ``dirbti, lipdavo'' may be mistakenly written as ``dirpti, libdavo'' as this is how they sound due to the assimilation.

\item Uppercasing or lowercasing the first letter in a word. For example, the word ``ąžuolas'' can start both with the lower or upper case depending on whether it is a tree (oak) or person's name. We exclude the first words in a sentence as these always have to start the upper case.

\item Delete and add space. We separately match all occurrences of spaces and all empty strings not at a word boundary.
\end{enumerate}

Some samples of the corrupted sentences are presented in Appendix \ref{appendix2}.

\subsection{Transformer Models}

In this work, we compare T5 \cite{stankevivcius2021generating} and ByT5 \cite{xue2021byt5} transformer models for grammatical error correction of Lithuanian. They are of sequence-to-sequence type. The encoder encodes the input sequence with attention operating on all input tokens while the decoder predicts output sequence tokens one by one, attending to tokens of both encoder (all) and decoder (only previous ones).

Below we further emphasize the properties of these models that make them appropriate for our task.

\subsubsection{T5}

The original T5 \cite{2020t5} was designed to be universal for multiple tasks. Authors showed that there is no difference whether a custom ``head'' is used (added on top of the pre-trained transformer) for fine-tuning purposes or a simple sequence-to-sequence formulation in text format is employed (no need to add additional weights to a pre-trained model). This way even tasks with outputs as float numbers can be formatted into a text-to-text format. Such generic task formulation made the T5 model very popular.

In previous work \cite{stankevivcius2021generating}, we adapted the T5 model for the generation of summaries of Lithuanian news articles. We trained a SentencePiece \cite{kudo2018sentencepiece} tokenizer on $10^{6}$ and the main model on 2\,027\,418 news articles. As a result, this model should be familiar with the Lithuanian language (both tokenizer and model weights) and we use it as the basis for our fine-tuning purposes.

\subsubsection{ByT5}

ByT5 is a follow-up model from the multilingual mT5 \cite{xue-etal-2021-mt5} and T5 \cite{2020t5}. The authors showed that adapting byte-level tokenization can lead to a much more efficient use of model parameters. As an example, the multilingual mT5 had over 66\% of its weights (for the base version) allocated to its multilingual word pieces (a total of 250\,000) related weights (input embedding matrix and decoder softmax layer) which were only sparsely updating during the training. Meanwhile, ByT5 vocabulary has only 384 items and the model reuses the saved parameters in more massive layers rather than indexing tokens. These benefits allowed ByT5 to surpass the small and the base versions of mT5 \cite{xue2021byt5}.

The introduction of finer byte-level tokenization is especially important for grammatical error correction. Typos, variants in spelling and capitalization, and morphological changes can lead to completely different sub-word tokens, while byte tokens are affected the least. The authors of ByT5 showed that their model outperforms mT5 if various types of noise are introduced. Therefore, we use this model in our study of Lithuanian grammatical error correction.

\subsection{Training Details}

To train the models, we used a GeForce RTX 2080 Ti GPU. Following the best practices with the T5 family of models \cite{2020t5, xue-etal-2021-mt5, rothe2021a, xue2021byt5}, we used the total batch size (number of samples to pass through the model before the gradient update) of 128 for fine-tuning. For ByT5 it was achieved by 128 gradient accumulation steps of batch size 1; while for T5, 64 gradient accumulation steps of batch size 2. We had to use multiple accumulation steps to process the total batch sequentially by smaller parts as the total batch did not fit into GPU memory at once. It took us approximately 100 hours for ByT5 and 30 hours for T5 fine-tuning. ByT5 took longer due to the longer sequences produced by finer byte-level tokenization.

We used the training script and Pytorch model implementation from the Hugging Face library \cite{wolf-etal-2020-transformers}. For simplicity, we employed an Adafactor optimizer \cite{pmlr-v80-shazeer18a} with a constant learning rate of 0.001. If not stated otherwise, we used all the default parameters as in the Hugging Face library version 4.12.0.

\subsection{Evaluation}

One of the most popular grammatical error correction evaluation metrics is ERRANT \cite{bryant-etal-2017-automatic}. It applies a set of rules operating over a set of linguistic annotations to construct the alignment and extract individual edits between corrupted, corrected, and gold-standard texts. This way precision, recall, and \textit{F}-score can be calculated. We customized the original ERRANT by using Hunspell dictionaries \cite{hunspell_lt}, stemmer\footnote{\url{https://pypi.org/project/PyStemmer/}}, spaCy version 3.2 pipeline \verb|lt_core_news_lg| \footnote{\url{https://spacy.io/models/lt}}, and corresponding part-of-speech tags for the Lithuanian language.

During the inference, we used simple greedy decoding with a beam size of 1. That is, we simply selected for each next token the one that the model assigned the highest probability to.

\section{Results}

\begin{figure}
\includegraphics[width=\textwidth]{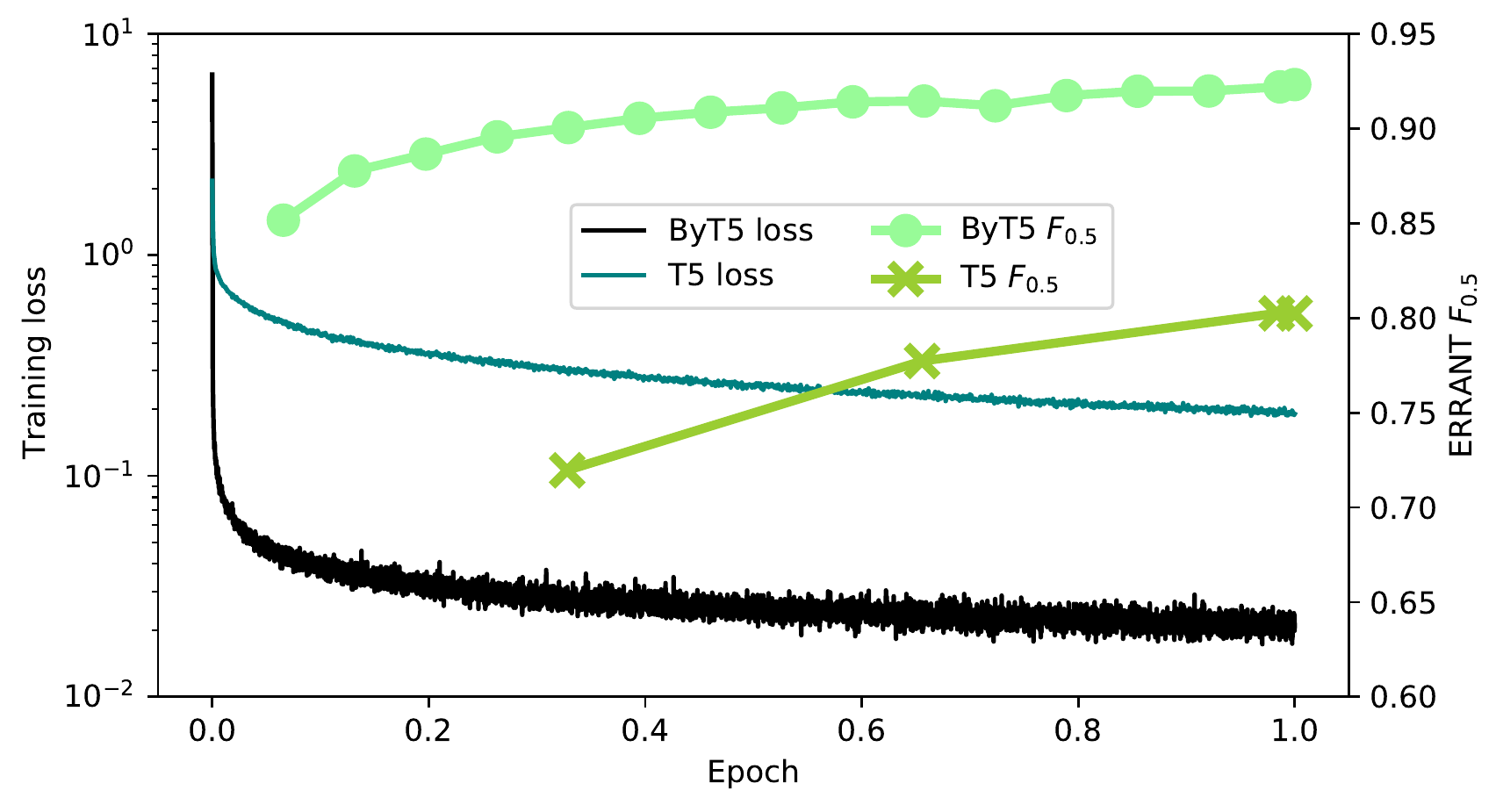}
\caption{Training loss and F$_{0.5}$ score for both T5 and ByT5 runs.}
\label{fig2}
\centering
\end{figure}

Training dynamics of T5 and ByT5 models are depicted in Figure \ref{fig2}. Only after 6\% of training, the ByT5 score F$_{0.5}=0.85$ is already higher than the F$_{0.5}=0.80$ T5 managed to reach after the full epoch. We can also see that the performance is steadily increasing during the fine-tuning and is expected to continue doing so. The same results are indicated by the training loss. It is much lower for the ByT5, hence the model is better.

We divided our synthesized errors into several groups and corrupted the test set with each group separately from the others. Evaluation results of such setup are presented in Table \ref{tab:groups}. We can see that the easiest task for both models was adding or deleting spaces, while the hardest task is correcting assimilation and gemination mistakes. This group may lag in performance due to the smaller abundance (2\% of samples) in the training data.

We present some generated test samples in Appendix \ref{appendix2}.

\begin{table}[h]
\setlength{\tabcolsep}{13.4pt}
  \caption{Evaluation for the separate error categories with models trained for one epoch. We applied synthetic corruption for the test set of ByT5 (total of 2\,155 samples) and T5 (total of 2\,099 samples) with each error group separately. We show both ERRANT F$_{0.5}$ score and number of samples (\#samples) affected and evaluated on.}
  \label{tab:groups}
\begin{tabular}{lrrrr}
\toprule
\multirow{2.5}{*}{Error group} &    \multicolumn{2}{c}{ByT5} &  \multicolumn{2}{c}{T5} \\
\cmidrule(r){2-3} \cmidrule(l){4-5}
& F$_{0.5}$ &\#samples & F$_{0.5}$ & \#samples\\
\midrule
Typographical             &  0.87 & 1\,916  &0.72 & 1\,868\\
Punctuation       &  0.81 &  489 &0.36 &460\\
Similar sounding letters      &  0.88 &  1\,115 &0.55 &1\,143\\
Add/delete spaces &  0.96 &  1\,873 &0.74 &1\,832\\
Assimilation/Gemination     &  0.79 & 56  &0.30 & 43\\
Upper/Lower casing          &  0.86 & 785  &0.47 &781\\
\bottomrule
\end{tabular}
\end{table}

\section{Discussion}

We trained the first reported deep-learning-based Lithuanian grammatical error correction system and compared two sequence-to-sequence transformer models for the task.

The ByT5 transformer model, based on byte-level tokenization, greatly outperformed the subword counterpart T5. We think that the main reason for this is that the fine-grained byte-level details allow the model to maximize acquired information about the sentence and thus calculate a more accurate representation. This way, the model sees a bigger picture and has to solve the task with less ambiguity. On the other hand, longer and more informative token sequences are slower to process and induce the slowdown of three times, compared to the T5. Yet even if we compare models trained for the same amount of time, ByT5 is still the leader. This shows that for the grammatical error correction it is crucial to have the best possible representation of the text.

We thought that during the T5 subword tokenizer training acquired common token patterns may be of great use. Yet our results show that this is not the case. On the contrary, it may make it harder for the model to ``understand'' the true representation behind the corrupted text.

In the future, we plan to train the ByT5 model even longer. It is clearly visible from our results that in the current state it is under-trained. Additional benefits could be expected from more data and more passes through the dataset.

We hope that this work will help both researchers and Lithuanian language users. We make our trained model and code available at \url{https://github.com/LukasStankevicius/Towards-Lithuanian-Grammatical-Error-Correction}.
 
\subsection*{Funding}

The research is partially funded by the joint Kaunas University of Technology Research and Innovation Fund and Vytautas Magnus University project ``Deep-Learning-Based Automatic Lithuanian Text Editor (Lituanistas)'', Project no.: PP34/2108.

\subsection*{Acknowledgements}

We thank our project collaborators from Vytautas Magnus University, especially Jurgita Kapočiūtė-Dzikienė, for valuable discussions on related topics.
 
\bibliographystyle{splncs04}
\bibliography{references}

\newpage
\appendix{}

\section{Statistics for corrupting similar letters and punctuation}\label{appendix1}

\begin{table}[h]
\setlength{\tabcolsep}{5.1pt}
\centering
  \caption{Regex expressions to find specific patterns in texts and statistics of distinct finds used as probability weights for replacement.}
  \label{tab:mixing_statistics}
\begin{tabular}{lrrrrrr}
\toprule
Group (regular expression) & \multicolumn{6}{c}{Matches and counts} \\
\midrule
\verb*@[,\.–]{0,1} @ & \verb*@ @ &  79\,695\,056 &   \verb*@. @ &   5\,125\,941 &  &\\
                        & \verb*@, @ &   9\,876\,726 &   \verb*@– @ &   1\,347\,515 & &\\
                        \midrule

\verb*@[\.,;:\–\-?!\(\)\[\]\<\>/]@ & \verb*@,@ &  10\,072\,919 &    \verb*@?@ &    300\,962  & \verb*@]@ &     34\,283\\
                        & \verb*@.@ &   7\,976\,435 &    \verb*@:@ &    519\,928 & \verb*@>@ &      5\,759 \\
                        & \verb*@–@ &   1\,453\,095 &    \verb*@!@ &    106\,333  & \verb*@<@ &     4\,457\\
                        & \verb*@)@ &    665\,253 &    \verb*@;@ &    105\,526 & &\\
                        & \verb*@(@ &    655\,651 &    \verb*@/@ &     90\,778 & &\\
                        & \verb*@-@ &    546\,698 &    \verb*@[@ &     34\,295 & &\\
\midrule
\verb*@u{0,1}ou{0,1}@ & \verb*@o@ &  33\,058\,916 &   \verb*@ou@ &     41\,509  & &\\
                        & \verb*@uo@ &   3\,355\,463 &  \verb*@uou@ &        34  & &\\
\midrule
\verb*@ia|e@ & \verb*@ia@ &   6\,733\,731 &    \verb*@e@ &  35\,509\,427  & &\\
\verb*@[scz]@ & \verb*@s@ &  47\,349\,069 &    \verb*@c@ &   2\,645\,328  & \verb*@z@ & 1\,646\,823\\
\verb*@[ščž]@ & \verb*@š@ &   7\,002\,598 &    \verb*@č@ &   2\,619\,317  & \verb*@ž@ & 5\,044\,500\\
\verb*@[eęė]@ & \verb*@e@ &  35\,509\,427 &    \verb*@ę@ &   1\,336\,170  & \verb*@ė@ & 9\,781\,460\\
\verb*@[iįy]@ & \verb*@į@ &   3\,490\,952 &    \verb*@y@ &   8\,347\,510  & \verb*@i@ & 82\,431\,807\\
\verb*@[uųū]@ & \verb*@ū@ &   2\,795\,974 &    \verb*@ų@ &   7\,826\,828  & \verb*@u@ & 28\,978\,236\\
\verb*@[aą]@ & \verb*@a@ &  68\,291\,558 &    \verb*@ą@ &   4\,471\,872  & &\\
\verb*@[cč]@ & \verb*@c@ &   2\,645\,328 &    \verb*@č@ &   2\,619\,317  & &\\
\verb*@[zž]@ & \verb*@z@ &   1\,646\,823 &    \verb*@ž@ &   5\,044\,500  & &\\
\verb*@[td]@ & \verb*@t@ &  35\,864\,854 &    \verb*@d@ &  14\,822\,144  & &\\
\verb*@[kg]@ & \verb*@k@ &  26\,461\,947 &    \verb*@g@ &  10\,626\,341  & &\\
\verb*@[pb]@ & \verb*@p@ &  16\,187\,509 &    \verb*@b@ &   8\,148\,725  & &\\
\midrule
\verb*@‘‘|,,|[„“"”]|''@ & \verb*@"@ &    436\,378 &   \verb*@,,@ &     11\,777  & \verb*@''@ &87\\
                        & \verb*@”@ &     46\,847 &   \verb*@‘‘@ &       817  & &\\
\bottomrule
\end{tabular}
\end{table}

\newpage
\section{Corruption and correction examples}\label{appendix2}

\begin{table}[h]
\setlength{\tabcolsep}{2.2pt}
\centering
  \caption{Samples of the original, corrupted, and corrected text forms. Here fully-trained (1 epoch) ByT5 models were used.}
  \label{tab:corurption_examples}
\begin{tabular}{ll}
\toprule
Type           &  Text\\
\midrule
Original & „Mes nenorime, kad jie keiktųsi, pyktųsi. Neleidžiame ne tik gerti, bet ir\\
&rūkyti. Taisyklės čia griežtos, rūkei, atleisime tau kartą, nepaklusai, eik\\
&iš kur atėjęs. Jei jau žmogus nusprendė keisti gyvenimą, tai turi būti\\
&daroma rimtai“, - nuolaidų nežada M. Balčiūnas.\vspace{5pt}\\
Corrupted & "Mes nenorime, kad jie keiktųsi, byktųsi. Nleeidžiame ne tik gerti, bet ir \\
&rūkyti. Taisyklės čia griežtos, Rūkei, atle isime tau kartą, nepaklusai, eik\\
&iš kur atėjęs.  Jei j au žmogus nuspr-endė keisti gyvenimą tai turi būti\\
&daromo rymtai“, - nuolaidų nežada M. Balčiūnas]\vspace{5pt}\\
ByT5&„Mes nenorime, kad jie keiktųsi, pyktųsi. Neleidžiame ne tik gerti, bet ir\\
&rūkyti. Taisyklės čia griežtos. Rūkei, atleisime tau kartą, nepaklusni, eik\\
&iš kur atėjęs. Jei jau žmogus nusprendė keisti gyvenimą, tai turi būti\\
&daroma rimtai“, - nuolaidų nežada M. Balčiūnas.\\
\midrule
Original & Šeštadienio vakarą Klaipėdoje surengto „Eurovizijos“ atrankos finalo\\
&dalyviai po renginio miegoti nėjo – dešimt savaičių trukusios kovos\\ 
& pabaigą atšventė uostamiesčio kokteilių bare „Oscar“.\vspace{5pt}\\
Corrupted & Šeštdienio vakarą Klaipėdoje surengto „Eurovizijos“ atrankos finalo\\
&dalyvia i po rengicio miegoti nėjo – dešimt savaičių drukusio kovos\\
&pabaigą atšventė uostamiesčio kokteilių bare „oscar“.\vspace{5pt}\\
ByT5 & Šeštadienio vakarą Klaipėdoje surengto „Eurovizijos“ atrankos finalo\\
&dalyviai po renginio miegotinėjo – dešimt savaičių trukusio kovos\\
&pabaigą atšventė uostamiesčio kokteilių bare „Roscar“.\\
\midrule
Original&300 kg hašišo gabenimo į Lietuvą byla: vienas išteisintas, kitam\\
&sušvelninta bausmė\vspace{5pt}\\
Corrupted&300 kg haši šo gabenimo į Lietuvą byla. vie nas i štsisintas, kitam\\
&sušverlninta buasmė\vspace{5pt}\\
T5&300 kg hašišo gabenimo į Lietuvą byla: vienas išteisintas, kitam\\
&sušvelninta bausmė\\
\bottomrule
\end{tabular}
\end{table}

\end{document}